# Planetary UAV localization based on Multi-modal Registration with Pre-existing Digital Terrain Model

*Xue Wan [a,b], Yuanbin Shao [a,b,c], Shengyang Li [a,b]*

[a] Technology and Engineering Center for Space Utilization, Chinese Academy of Sciences, Beijing, China; wanxue@csu.ac.cn
[b] Key Laboratory of Space Utilization, Chinese Academy of Sciences, Beijing, China;
[c] School of Aeronautics and Astronautics, University of Chinese Academy of Sciences, Beijing, China;

**Abstract:**

The autonomous real-time optical navigation of planetary UAV is of the key technologies to ensure the success of the exploration. In such a GPS denied environment, vision-based localization is an optimal approach. In this paper, we proposed a multi-modal registration based SLAM algorithm, which estimates the location of a planet UAV using a nadir view camera on the UAV compared with pre-existing digital terrain model. To overcome the scale and appearance difference between on-board UAV images and pre-installed digital terrain model, a theoretical model is proposed to prove that topographic features of UAV image and DEM can be correlated in frequency domain via cross power spectrum. To provide the six-DOF of the UAV, we also developed an optimization approach which fuses the geo-referencing result into a SLAM system via LBA (Local Bundle Adjustment) to achieve robust and accurate vision-based navigation even in featureless planetary areas. To test the robustness and effectiveness of the proposed localization algorithm, a new cross-source drone-based localization dataset for planetary exploration is proposed. The proposed dataset includes 40200 synthetic drone images taken from nine planetary scenes with related DEM query images. Comparison experiments carried out demonstrate that over the flight distance of 33.8km, the proposed method achieved average localization error of 0.45 meters, compared to 1.31 meters by ORB-SLAM, with the processing speed of 12hz which will ensure a real-time performance. We will make our datasets available to encourage further work on this promising topic.

**Key words:**

UAV Localization; Multi-modal registration; Frequency; SLAM

## 1. Introduction

Planetary exploration is important to the answers to many fundamental questions including the formation of the universe, the evolution of life and the origin of the Earth. UAVs can reach places and areas where man is unable or very difficult to access. For planetary exploration, UAVs can be the effective tool for investigation and survey. NASA (National Aeronautics and Space Administration) has led a project titled "BEES for Mars", which utilizes UAVs for Mars exploration (Plice et al., 2003). Ingenuity, the Mars helicopter (NASA, 2020), has succeeded in landing on Mars surface together with Perseverance rover and flied for six times already. NASA also planned to send a mobile robotic rotorcraft named as Dragonfly to Titan, the largest moon of Saturn (E.P. et al., 2017).

As one of the key technologies in planetary exploration, robust and accurate navigation system ensure the safety and precise of the scientific tasks. Unlike on Earth, GPS service is unavailable for planetary exploration, and thus, optical based navigation has become one of the key technologies in autonomous navigation (Matthies et al., 2007). There are several benefits in optical based navigation including low power consumption, light weight and small size. Optical sensors can be used for multiple purposes, including obstacle avoidance (Janschek et al., 2006), target area detection (Matthies et al., 2007), 3D terrain surface reconstruction (Cheng, Maimone, et al., 2005), etc.

As there is hardly any artificial architecture on planetary surface, terrain feature becomes one of the fundamental and crucial processing unit in vision-based navigation on planetary surface. To utilize terrain feature for planetary navigation, several approaches have been proposed, including skyline matching (Cozman et al., 2000) (Chiodini et al., 2017), ray tracing (Fang, 2020), 3D topography matching (A. V. Nefian et al., 2014), etc. However, none of these methods are able to solve the multi-modal matching problem between topography and optical imagery, in this paper, we aimed to answer the question: "How can we find the shared features between digital terrain model and optical UAV imagery"?

Directly match the optical images and DEM data have following merits: firstly, DEM data can be acquired from stereo-matching or SAR interference and is always available for planetary exploration. Secondly, as terrain feature is important in planetary navigation, DEM data is the representation of terrain model, so terrain features can be directly extracted and matched. Thirdly, the illumination effect between UAV and satellite can be largely eliminated. The reference map, terrain shaded images, can be generated under the similar illumination condition as UAV images.

In this paper, we propose a terrain-aided navigation approach based on frequency domain phase correlation which is able to localize the UAV in a terrain elevation map. The global localization results will be fused with visual odometry estimation results via LBA by proposing a new cost function, as illustrated in Figure 1.

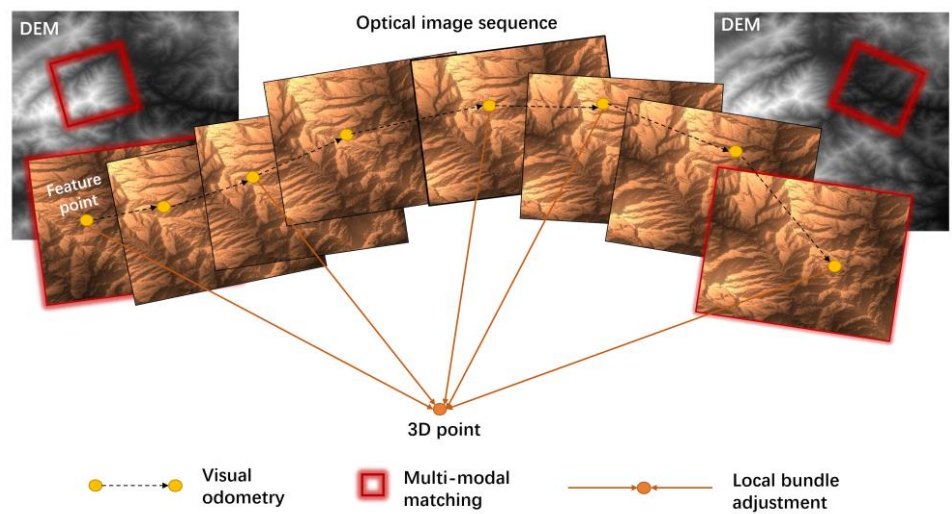

Figure 1 Geo-referencing based optical navigation for planetary UAV navigation

The contributions of this paper are summarized as:

1) This paper proposed a new planetary UAV localization dataset including 40200 synthetic drone images with ground truth terrain model taken under nine different planetary topography.

2) We proved via mathematical derivation that the Fourier spectrum of optical images can be divided into illumination and topography features, and the topography features of terrain model and optical imagery can be correlated via frequency-based correlation.

3) This paper propose terrain aided optical navigation algorithm, terrain-SLAM, which combined odometry estimation and frequency-based geo-referencing via a newly designed local bundle adjustment framework.

## 2. Related work

One of the key challenges in planetary UAV optical navigation is to find a suitable image matching/image retrieval method for multi-modal planetary images. All the approaches can be divided into five categories: feature-based matching, area-based matching, skyline matching, 2D-3D ray tracing and 3D topography matching.

Feature matching is one of the most commonly used matching algorithm for planetary optical navigation. FP (Fixed Point) feature (Misu et al., 1999) and Shi-Tomasi feature based image matching (Johnson et al., 2000) has been applied in Itokawa asteroid exploration. Harris corner detection and feature matching are used for Mars rover (Cheng, Johnson, et al., 2005), Curiosity and Opportunity, in the landing process. SIFT feature matching has been applied in the CE-3 landing on the moon (Z. Liu et al., 2014). Recently, deep learning based matching, such as SuperPoint (Detone et al., 2018), has been introduced in image matching. Characteristics such as self-supervised training, homographic adaptation, and cross-domain adaptation in SuperPoint have made great progress in terms of efficiency and accuracy in visual SLAM for robots. Feature-based matching, which has been improved for years in the computer vision society, is able to cope with relatively large geometric image distortions between images. However, they can hardly cope with featureless areas which commonly exist in planetary surface.

Another type of matching algorithm is area-based or correlation-based algorithms. Typical area-based matching algorithms include NCC (Normalized Cross Correlation), MI (Mutual Information)(Papoulis & Pillai, 2002) and LSS (Local Self-Similarities) (Shechtman & Irani, 2007). SAD (Sum of Absolute Difference) has been used for Mars rover obstacle avoidance, and further improved by Chilian and Hirschmüller (Chilian & Hirschmüller, 2009) using five matching windows for the task of Mars terrain 3D reconstruction. Compared to feature-based algorithms, area-based matching algorithms are more robust to featureless area as they take global grey value distribution. However, they are sensitive to camera pose variation which make them mostly applied in stereo-vision instead of optical navigation in planetary exploration.

Feature-based approaches generally try to extract the high frequency information, corners, edges, etc., from the planetary images, which are not commonly existed in featureless natural scenes and can be altered greatly by the change of illumination conditions. Another type of solution is to use topography features, as they are distinctive and unique and largely exist even in featureless areas, and thus, DEM (Digital Elevation Model) is widely used as reference map in for absolute localization for planetary exploration.

The most common DEM aided navigation is skyline matching (Cozman et al., 2000) (Chiodini et al., 2017). These approaches first extract a skyline from the rover image, and then a simulated skyline is rendered at an estimated rover position in the digital elevation model. Finally, similarity values between the extracted skylines and the simulated skylines are calculated. The high similarity value stands for high likelihood of the estimated rover position. Although skyline matching are commonly used to localize a rover within a known digital elevation model, it is not suitable for planetary UAV navigation as the view angle of UAV is nadir and the skyline may not be visible. Moreover, the accuracy of skyline matching is largely determined by the DEM resolution, that may lead to localization errors if the resolution of DEM is not high enough.

Another type of terrain relative navigation approach is 3D topography registration between the DEM and the generated point cloud from on-board sensors, such as cameras and LiDAR. While dense 3D point cloud can be directly generated from LiDAR scanning, the high energy requirement prevent it from widely utilization in planetary navigation. 3D point cloud can also be generated from stereo matching using the algorithms in Table 1, however, they may suffer from the featureless areas. Even the 3D point cloud can be successfully generated, how to register it with DEM is another problem. One of the commonly used 3D point cloud registration algorithm is ICP(Iterative Closest Point) (A. V. Nefian et al., 2014) (Ara V Nefian et al., 2017), but ICP can hardly work with occlusion when a rover is facing a small hill. To overcome the limitation of ICP, Fang (Fang, 2020) proposed a ray tracing approach to register a stereo imagery to lunar terrain model. This approach however, still suffer from variations between the generated 3D point cloud and DEM owing to the largely different viewing distance and angles.

The merits and disadvantages of the five state-of-the-art planetary navigation approaches are summarized in Table 1. As shown in Table 1, none of the approaches are able to cope with multi-modal difference between terrain model and optical UAV images.

Table 1 Summary of state-of-the-art planetary navigation approaches

| Methods | Perspective robustness | Scale robustness | Illumination robustness | Multi-modal robustness | Computational efficient |
|---|---|---|---|---|---|
| Feature-based | ✓ | ✓ | ✗ | ✗ | ✗ |
| Area-based | ✗ | ✓ | ✗ | ✗ | ✗ |
| Skyline matching | ✓ | ✓ | ✗ | ✗ | ✓ |
| Ray tracing | ✓ | ✗ | ✓ | ✗ | ✓ |
| 3D topography matching | ✓ | ✗ | ✓ | ✗ | ✗ |

In this paper, we propose to use frequency information for multi-modal matching in order to localize the current view of UAV in a digital elevation map. To the best of our knowledge, none of the above approaches try to directly match the optical image with the DEM. This may because the features extracted from optical image and DEM are so different that it becomes really difficult to correlate them. In this paper, we demonstrated that although it is visually difficult to locate the optical image in DEM, their frequency information can be correlate via phase correlation and thus, the multi-modal image matching can be done in frequency domain.

Then, the next question goes to how to iteratively estimate the six DOF of the planetary UAV. In this paper, we seamlessly incorporate multi-modal registration into a LBA (Local Bundle Adjustment) algorithm by propose to use the phase information to correlate the topography features between DEM data and optical image. Finally, we proposed a terrain aided SLAM is proposed by seamlessly integrating the multi-modal registration result into a visual odometry estimation.

# 3. Theoretical analysis for frequency-based terrain feature matching

## 3.1 Phase Correlation

This section will briefly introduce the principle and history of phase correlation, a frequency-based image matching algorithm which estimates the image shift between two images via shift property of Fourier Transform (FT). According to the shift property of FT, the translation shift $(a, b)$ in the image spatial domain results in a linear phase shift in the Fourier frequency domain. Then, phase correlation is defined as the normalized cross power spectrum $Q(u, v)$ between the Fourier transforms of the two images $I_1(x, y)$ and $I_2(x, y)$.

$$Q(u, v) = \frac{F_1(u, v) F_2^*(u, v)}{|F_1(u, v) F_2^*(u, v)|} \qquad (1)$$

where * stands for complex conjugate. $F_1(u, v)$ and $F_2(u, v)$ in $(u, v)$ coordinates are the Fourier transforms of the two images $I_1(x, y)$ and $I_2(x, y)$.

The shifts $(a, b)$ can be resolved at integer level via IFT (Inverse Fourier Transform) to convert $Q(u, v)$ to an approximate Dirac delta function as

$$IFT\big(Q(u, v)\big) = \delta(x - a, y - b) \qquad (2)$$

The translation $(a, b)$ can also be solved directly in frequency domain with sub-pixel accuracy by unwrapping and fitting the PC fringes in the cross power spectrum $Q(u, v)$ (Balci & Foroosh, 2006) (Hoge, 2003)(J. G. Liu & Yan, 2008) or in spatial domain by fitting the peak of function $\delta$ to a Gaussian function (Argyriou & Vlachos, 2004). The peak value of function $\delta$, ranging from 0 to 1, indicates the quality of PC matching. If the two images are identical, the peak value of $\delta$ equals 1.

In our previous work, we have proved the robustness of Phase Correlation to sun angle variation and shadow via PCD (Phase Correlation Decomposition) model

(Wan et al., 2015). Using the illumination invariant property of PC, we applied it to illumination invariant change detection based on pixel-pixel matching (Wan, Liu, et al., 2019) and UAV navigation based on the geo-referencing between UAV image and reference satellite images (Wan et al., 2016). To enhance the robustness of phase correlation towards complex geometric distortion, we combine phase correlation with particle swarm optimization (Wan, Wang, et al., 2019). We demonstrate the superior performance of phase correlation towards state-of-the art image matching algorithms in UAV navigation tasks (Wan, Wang, et al., 2019). In this paper, we will further investigate the robustness of PC to multi-modal data including DEM and optical imagery via theoretical analysis. While our previous work focused on the image matching algorithm, this paper will propose a fully terrain aided navigation pipeline which combine geo-referencing results and visual odometry results for planetary UAV navigation.

**3.2 The robustness of Phase Correlation for DEM based geo-referencing**

In this section, we will dig out the relationship between UAV optical image and the topography model in frequency domain, and then proved via mathematical derivation that the terrain feature shared by optical image and the topography model can be extracted and correlated via phase correlation.

Firstly, we will figure out the mathematical relationship between optical image UAV image $I(x,y)$ and the topography model $V_H(x,y)$ in spatial domain. As planetary surface is natural landscape, in this paper, it is regarded as a Lambertian model. Under a given solar radiation intensity $L$ in direction of azimuth angle $\tau$ and elevation angle $\sigma$, the intensity of UAV image $I(x,y)$ generated from elevation model $V_H(x,y)$ is (Kube & Pentland, 1988)

$$I(x,y) = L \frac{p\cos\tau\cos\sigma + q\sin\tau\cos\sigma + \sin\sigma}{\sqrt{(p^2 + q^2 + 1)}} r(x,y) \qquad (3)$$

where $p = \frac{\partial}{\partial x} V_H(x,y)$, $q = \frac{\partial}{\partial y} V_H(x,y)$ are the gradients of $V_H(x,y)$ in $x$ and $y$ direction. $r(x,y)$ represents the reflectance value at position $(x,y)$.

To analysis their relationship in frequency domain, the UAV image $I(x,y)$ is transformed into frequency domain as

$$F_I(\omega,\theta) = L[\sin\sigma + \cos\sigma\cos(\theta - \tau) F_H(\omega,\theta)] \tag{4}$$

where $F_I(\omega,\theta)$ is the Fourier spectrum of elevation $I(x,y)$, $F_H(\omega,\theta)$ is the Fourier spectrum of elevation $V_H(x,y)$.

It should be noted that here we use a polar coordinate instead of Cartesian coordinate to better illustrate the illumination effect on frequency domain.

Then, the PC cross power spectrum of optical image $F_I(\omega,\theta)$ and DEM $F_H(\omega,\theta)$ is:

$$Q(\omega,\theta) = \frac{\cos\sigma}{Z_d}\cos(\theta - \tau) + \frac{\sin\sigma}{Z_d}F_H^*(\omega,\theta) \tag{5}$$

where $Z_d = |\cos\sigma F_H^*(\omega,\theta) + \sin\sigma\cos(\theta - \tau)|$.

As shown in Equation (6), the cross power spectrum of an UAV optical image and reference DEM $Q(\omega,\theta)$ is mainly altered by two terms, the illumination term, $\cos(\theta - \tau)$ and the topography term, $F_H^*(\omega,\theta)$. The following sections will investigate the effect of each term respectively.

### 3.2.1 The effect of illumination term

To analysis the effect of illumination term, $\cos(\theta - \tau)$, we assume that a UAV image $I(x,y)$ shown in Figure 2(a) is generated under a small elevation angle $\sigma = 10°$. In this case, term $\sin\sigma\cos(\theta - \tau)$ becomes dominant in PC cross power spectrum $Q(\omega,\theta)$ with other terms are nearly zero, and then $Q(\omega,\theta)$, shown in Figure 2(c), can be simplified as

$$Q(\omega,\theta) = \frac{\cos\sigma}{Z_d}\cos(\theta - \tau) \tag{6}$$

Then, we can consider that there is a shift between the $I(x,y)$ and $V_H(x,y)$, so their cross power spectrum, shown in Figure 2(d), becomes

$$Q(\omega,\theta) = \frac{\cos\sigma}{Z_d}\cos(\theta - \tau) e^{-i(au+bv)} \tag{7}$$

As shown in Figure 2(d), several fringe patterns are flipped owing to the term $\cos(\theta - \tau)$. If the value of $\cos(\theta - \tau)$ is negative, the fringes will be flipped. However, for both positive and negative correlation in the PC cross power spectrum remains the same and thus the fringe density $\sqrt{(a^2 + b^2)}$ and orientation *b/a* relating to the translation remains the same.

As the $\sin\sigma$, $Z_d$ and $\cos(\theta - \tau)$ are all real number, the inverse Fourier

transform of $Q(\omega, \theta)$ is still an approximate to Dirac delta as shown in Figure 2(e)

$$IFT(Q(\omega, \theta)) = \delta(x - a, y - b) \tag{8}$$

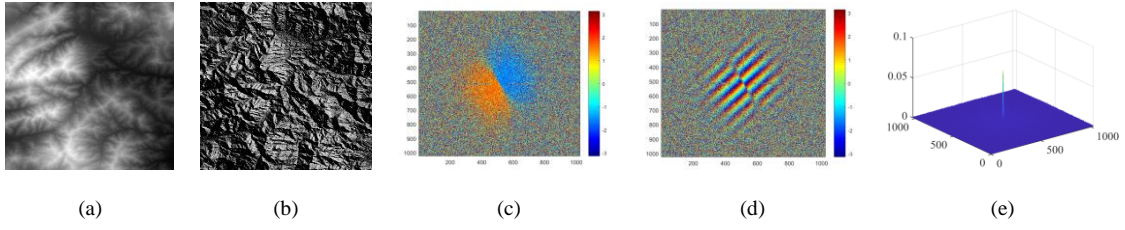

(a) (b) (c) (d) (e)

Figure 2 Image pair and their phase correlation result. Figure (a) is DEM image and (b) is optical image generated under illumination of $\sigma = 10°, \tau = 60°$. Figure(c) and (d) are PC cross power spectrum $Q(\omega, \theta)$ without and with image shift, and (e) is the inverse Fourier transform of $Q(\omega, \theta)$.

### 3.2.2 The effect of topography dependent term

When elevation angles $\sigma$ are large, the value of $\cos \sigma$ is close to 0 and the term $\sin\sigma\cos(\theta - \tau) \approx 0$, so the PC cross power spectrum $Q(\omega, \theta)$, shown in Figure 3(c), can be expressed as

$$Q(\omega, \theta) = \frac{F_H^*(\omega, \theta)}{Z_d} \tag{9}$$

Here $Q(\omega, \theta)$ becomes *illumination independent*, which means that the phase angle in PC cross power spectrum is determined only by topography, while irrelevant of illumination condition.

The cross power spectrum of $I(x, y)$ and $V_H(x, y)$ with image shift is

$$Q(\omega, \theta) = \frac{F_H^*(\omega, \theta)}{Z_d} e^{-i(au+bv)} \tag{10}$$

Different from the effect of *illumination term*, the effect of *topography* term is not a systematically flipping of the fringe patterns, but a randomly flipping of the fringes as shown in Figure 3(d). For a completely flat area, $V_H(x, y)$ is a constant and its Fourier transform $F_H^*(\omega, \theta)$ does not affect the phase correlation Fourier spectrum $Q(\omega, \theta)$. For areas with topographic relief, the topographic vectors, $\frac{\partial}{\partial x} V_H(x, y)$ and $\frac{\partial}{\partial y} V_H(x, y)$, are variables depending on slope, the results of $F_H^*(\omega, \theta)$ vary with topography. In other words, the impact of *topography term* degrades the power spectrum $Q(\omega, \theta)$, however, the fringe density and orientation relating to the translation remains the same.

The inverse Fourier transform of $Q(\omega, \theta)$ is an approximation of Dirac delta

function, as shown in Figure 3(e)

$$IFT(Q(\omega,\theta)) = \rho_H \delta(x-a, y-b) \qquad (11)$$

where $\rho_H$ is determined by the term of $FT(\frac{F_H^*(\omega,\theta)}{Z_d})$.

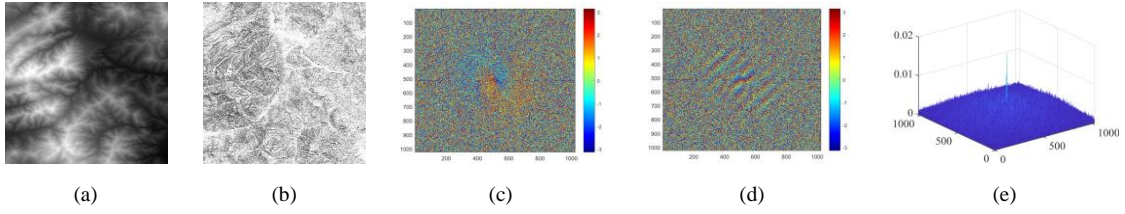

(a)      (b)      (c)      (d)      (e)

Figure 3 Image pair and their phase correlation result. Figure (a) is DEM image and (b) is optical image generated under illumination of $\sigma = 5°, \tau = 60°$. Figure(c) and (d) are PC cross power spectrum $Q(\omega,\theta)$ without and with image shift, and (e) is the inverse Fourier transform of $Q(\omega,\theta)$.

In general case, the negative correlation and randomly flipping introduced by *topography dependent* term and *illumination dependent* term will co-exist in PC cross power spectrum. However, none of the terms generate $2\pi$ wrapped periodic fringe patterns as introduced by image translation, $e^{-i(au+bv)}$. Therefore, although DEM data and optical image have large difference in spatial domain, their topography similarity can be identified in Frequency domain. This is the key property of PC based image matching to make it robust to multi-modal image registration.

Figure 4 shows the PC cross power spectrums generated from the five image pairs, one optical image and one DEM image, under different illumination conditions. It can be seen from the five figures that the whole cross power spectrums are divided into two angular parts: positive correlation parts shown by orange color and the negative correlation pars shown by blue color. This verified the Equation (13), that the value of $Q(\omega,\theta)$ is determined by the term $\cos(\theta-\tau)$. If the value of $\cos(\theta-\tau)$ is negative, the cross power spectrum will become negative. As shown in Figure 4, under the same azimuth angle $\tau$, the dividing line between positive correlation and negative correlation remains the same.

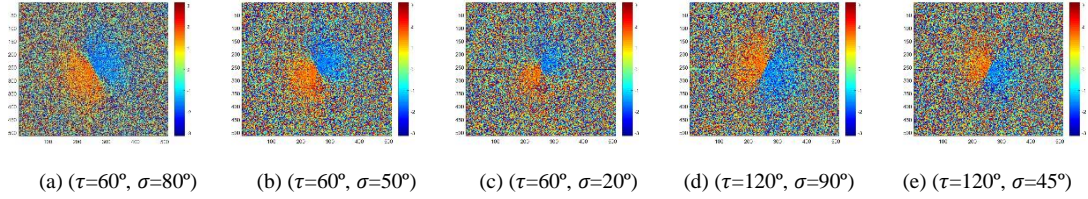

(a) (τ=60°, σ=80°)   (b) (τ=60°, σ=50°)   (c) (τ=60°, σ=20°)   (d) (τ=120°, σ=90°)   (e) (τ=120°, σ=45°)

Figure 4 Cross power spectrum maps and the from the image pairs. The caption below demonstrates different illumination conditions.

To further verified the relationship between $Q(\omega,\theta)$ and $\cos(\theta-\tau)$, the cross power spectrums $Q(\omega,\theta)$ are plotted with respect to azimuth angle $\tau$ ranges from $[-\pi,\pi]$, shown by the light blue in Figure 5(a)-(e). Meanwhile, the theoretical value of $Q(\omega,\theta)$ according to Equation (10) is also plotted to be compared with the experiment results.

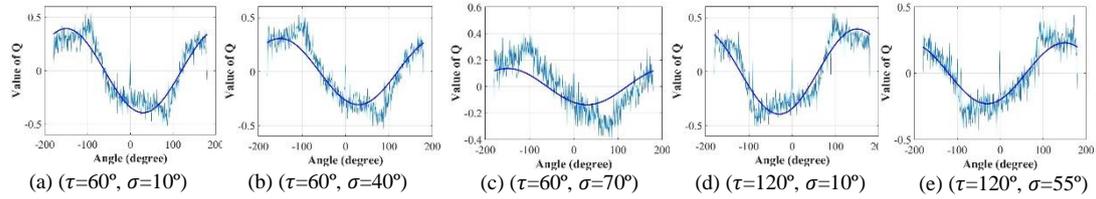

(a) (τ=60°, σ=10°)   (b) (τ=60°, σ=40°)   (c) (τ=60°, σ=70°)   (d) (τ=120°, σ=10°)   (e) (τ=120°, σ=55°)

Figure 5 Comparison between the $Q(\omega,\theta)$ generated from image matching experiments (light blue) and $\frac{\cos\sigma}{Z_d}\cos(\theta-\tau)$ (dark blue).

It can be concluded from Figure 5 that, generally, the $Q(\omega,\theta)$ values are in accordance with the illumination term $\cos(\theta-\tau)$, as the theoretical relationship between illumination condition and cross power spectrum derived in Equation (10) can be proved. There appears random noise in the experimental results of $Q(\omega,\theta)$, shown in light blue, and this is because the exists of topography terms. In fact, according to Equation (6), the PC cross power spectrum is a sum of both illumination term and topography terms weighted by elevation angle $\sigma$, and thus when elevation angles $\sigma$ are small, the illumination term is dominant and the values of $Q(\omega,\theta)$ are largely determined by the value of $\cos(\theta-\tau)$, as shown in Figure 5(a) and (d). With the increase of elevation angle $\sigma$, the topography term become dominant, and thus the effect of $\cos(\theta-\tau)$ becomes weak as shown in Figure 4 (c) that the blue curves become flat compared with other curves shown in Figure 4.

## 4. Terrain-aided Optical Navigation

Based on the robustness of Phase Correlation to multi-modal image matching, an terrain aided optical navigation approach is proposed and the main idea is demonstrated in Figure 6. Feature points are tracked frame-frame via visual odometry while key frames shown in red are geo-referenced to DEM data and then fused into a LBA module to generate to final camera poses and 3D point map.

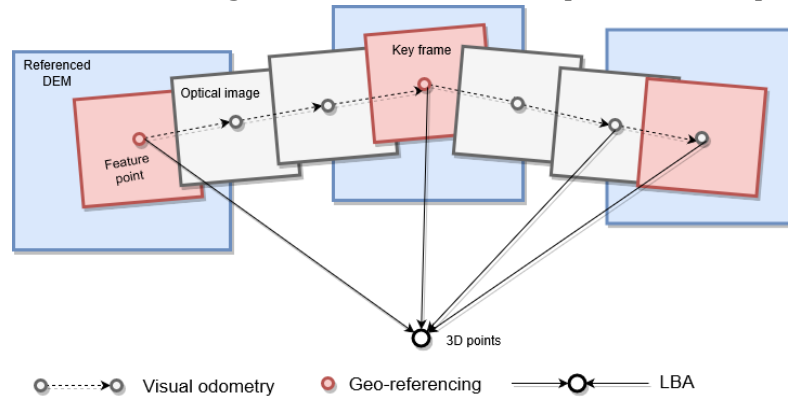

Figure 6 Overview of terrain-aided optical navigation for planetary exploration.

The pipeline of the proposed optical navigation approach is shown in Figure 7, which can be generally divided into three parts: visual odometry, geo-referencing and back-end optimization. Visual odometry estimated the 6 DOF camera pose via feature tracking. Then, key frames are selected and localized to reference map based on Phase Correlation image matching. The trajectories estimated by visual odometry and geo-referencing are aligned and fused into a back-end optimization. A new cost function is proposed for LBA algorithm which fuse the relative pose estimation from visual odometry and absolute pose estimation from geo-referencing to achieve optimal pose estimation.

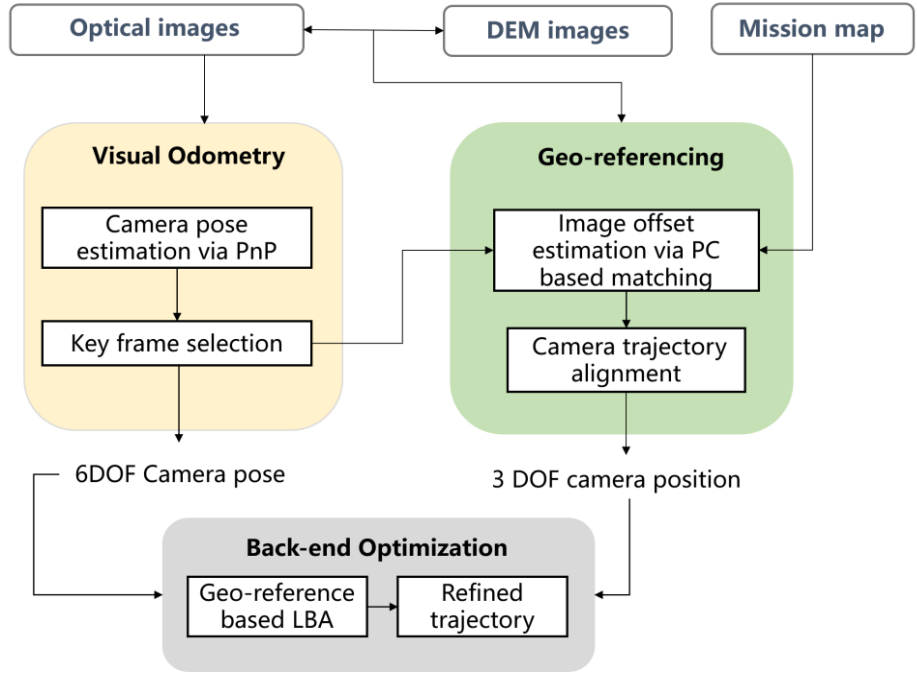

Figure 7 Pipeline of the proposed terrain aided geo-referencing for planetary UAV optical navigation approach

**4.2 Visual Odometry based on frame-frame feature tracking**

Visual odometry is carried out via frame-frame feature tracking. The 2D coordinate of feature points is determined via ORB features extraction and tracking (Mur-Artal et al., 2015). Let $m_k^i = [u^i, v^i]$ to be the $i^{th}$ coordinates of 2D feature points on the image $I_k$ at time $k$.

The initial 3D point $M_j = [x_j, y_j, z_j]$ can be reconstructed using triangulation from 2D corresponding points via frame-frame matching.

The relationship between 2D and 3D points can be obtained via projection function $\pi(\cdot)$:

$$m_k^i = \pi(P_k, M_j) \tag{12}$$

where $P_k \in SE(3)$ is the camera 3D pose of image $I_k$.

$$SE(3) = \left\{ T = \begin{bmatrix} R & t \\ 0^T & 1 \end{bmatrix} \in \mathbb{R}^{4 \times 4} \mid R \in \mathbb{R}^{3 \times 3}, t \in \mathbb{R}^3 \right\} \tag{13}$$

The camera pose $P_k$ can be estimated with known 3D map point $M_j$ and 2D feature point $m_k^i$ via EPnP.

Then, the camera trajectory $t_{vo}$ can be recovered defined as a serial of camera coordinate with the increase of time $k$.

$$t_{vo} = \{S_0^t, S_1^t, S_2^t, \cdots, S_k^t\} \tag{14}$$

where $S_k^t = [X_k^t, Y_k^t, Z_k^t]$ is calculated from camera pose $P_k$ as

$$S_k^t = S_{k-1}^t + t_k \tag{15}$$

In visual odometry, the camera trajectory $t_{vo}$ is in the local reference frame which the take the first camera frame position $S_0$ as coordinate origin. The visual odometry does not solve the problem of estimation rover's position in a global map coordinate system, which can be solved via geo-referencing.

As it is not computational efficient to take every image frame for geo-referencing, several key frames $I_k^i$ are selected for geo-referencing. The key frame selection is based on the following criterion: 1) the time interval two key frames should not be too small, otherwise, there is no point for key frame selection. 2) key frame are opt to be salient with rich features, so as to be matched with reference map.

Thus, in this paper, we adopted the key frame selection strategy as in ORB-SLAM (Mur-Artal et al., 2015):

1) More than 20 frames have passed from the last key frame;

2) There are more than 50 feature points in the current frame;

3) current frame tracks less than 90% points than reference keyframe.

**4.3 Phase Correlation for terrain based geo-referencing**

The main purpose of geo-referencing is to provide absolute position for image frames in global map coordinate system. As proved in section 3, Phase Correlation is able to correlate the topographic feature between multi-modal images. Another merit of Phase Correlation for absolute navigation is that PC is a non-iterative matching algorithm. This means that the image offset can be directly calculated without roaming searching even if target UAV images and the reference terrain shading images only share a small part of overlapping region. Thus, the PC based image matching can largely shorten the matching time which will ensure the real-time optical navigation.

The offset between the current UAV image frame $I_k$ and the reference terrain shading images $R_k$ can be obtained via PC based image matching

$$(\Delta X_k, \Delta Y_k) = PC(I_k, R_k) \tag{16}$$

The reference terrain shading images $R_k$ are generated using Equation (3) under presumably illumination condition of given solar radiation intensity $L$ in direction of azimuth angle $\tau$ and elevation angle $\sigma$. To speed up the image registration, the DEM data is cropped to local DEM according to the planned flight path $T_{pl}$

$$T_{pl} = \{S_0^R, S_1^R, S_2^R, \cdots, S_k^R | S_k^R = [X_k^R, Y_k^R, Z_k^R]\} \tag{17}$$

According to the offset $(\Delta X_k, \Delta Y_k)$ between current UAV images and the reference terrain shading images, the camera trajectory in global map coordinate system $T_{vo} = \{S_0^T, S_1^T, S_2^T, \cdots, S_k^T | S_k^T = [X_k^T, Y_k^T, Z_k^T]\}$ can be estimated via the following equation:

$$\begin{aligned} X_k^T &= X_k^R + GSD \cdot \Delta X_k \\ Y_k^T &= Y_k^R + GSD \cdot \Delta Y_k \\ Z_k^T &= DEM(X_k^R, Y_k^R) + H \end{aligned} \tag{18}$$

where $DEM(X_k^R, Y_k^R)$ is the elevation of the ground surface at $X_k^R$ and $Y_k^R$ coordinates, $H$ is the height of the UAV to the ground surface. $GSD$ is ground sample distance. Noted that the UAV fly relatively low at the Martian surface, the $GSD$ will change with the change of flight height and surface topography as shown in Figure 8.

$$GSD = 2\frac{Z_k^R - H_{min}}{s} tan\theta \tag{19}$$

where $H_{min}$ is the lowest altitude position in the field of view of the camera at this altitude, which is related to the focal length of the camera, and $s$ is the size of the UAV image defined as

$$s = \max(w, h) \tag{20}$$

where $w$ and $h$ is the width and height of the UAV images. For example, if the image size is 720×480 pixels, the value of $s$ is 720.

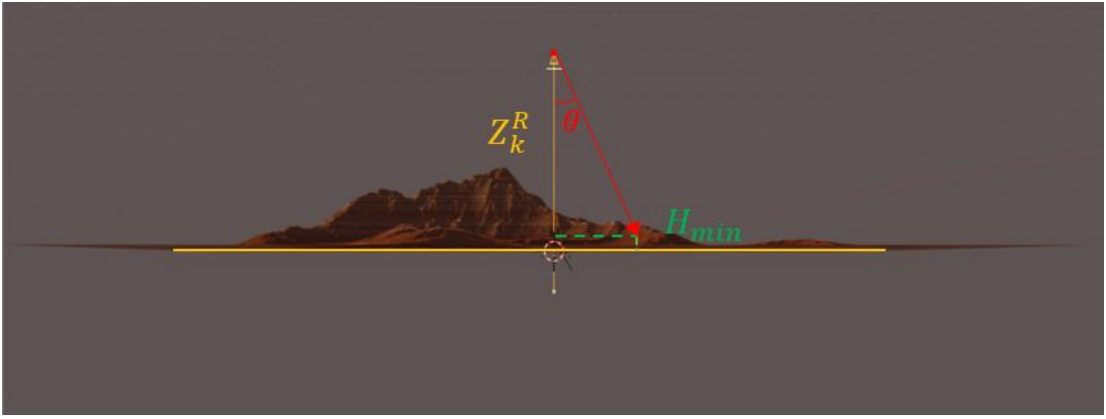

Figure 8 The illustration of *GSD* with respect to flight height and topography

From geo-referencing based on PC matching the camera trajectory $T_{vo}$ of the key frame is recovered in global map coordinate system, while in visual odometry the camera trajectory $t_{vo}$ is estimated in local reference coordinate system. The two trajectories are required to be transformed into the same coordinate system in order to be fused into a Local Bundle Adjustment. The transformation between the two trajectories can be calculated as

$$T_{vo} = H_{Tt} t_{vo} \tag{21}$$

where $H_{Tt} \in SE(3)$.

**4.3 Key frame pose refinement based on Local Bundle Adjustment**

The results of visual odometry and geo-referencing are complementary. The error accumulation or 'trajectory drift' can be corrected by the result of geo-referencing via global matching, while visual odometry can produce 6 DOF camera poses compared to 3 DOF camera pose estimated from geo-referencing. Thus, LBA is performed to fuse the visual odometry and geo-referencing results into more reliable and accurate camera poses.

BA (Bundle adjustment) can be defined as the problem of simultaneously refining the 3D coordinates describing the scene geometry, the parameters of the relative motion, and the optical characteristics of the camera(s) employed to acquire the images, according to an optimality criterion involving the corresponding image projections of all points (Triggs et al., 2000). Instead of taking all 3D point and camera poses for optimization, which is time-consuming, LBA takes the several adjacent camera positions, such as 15 frames, for optimization. This will largely increase the optimization speed during real-time navigation.

LBA can be done for fixed interval, such as every 15 or 20 image frames. The problem is that the error propagation in SLAM is not a linearly increase with the distance. For example, the error will propagate faster when UAV is turning or changing direction than keeping straight forward. There might also appear large localization errors when few features can be extracted from optical images.

Traditionally, LBA is achieved by minimizing the reprojection error between the image locations of observed and predicted image points. Figure 9(a) shows the network structure of four 3D points 1-4 and three images A-C. 3D point 1 are seen in image A and B, point 2 seen in A, B and C, point 3 seen in A and C, while point 4 seen in B and C. Then, the cost function $f(x)$ is defined as

$$f(x) = \frac{1}{2}\sum_{i=0}^{M}\sum_{j=0}^{N} e_{i,j}^T w_i e_{i,j} \tag{22}$$

where $e_{i,j} = \pi(P_k, M_j) - m_k^i$ is the reprojection error of 3D point $M_j$ on the image $I_i$ and $w_i$ is the confidence of observation $m_k^i$.

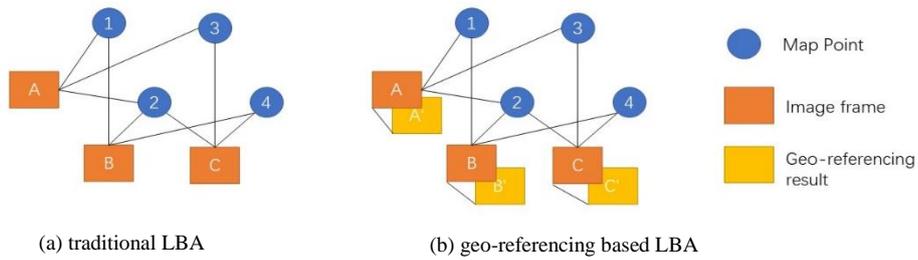

(a) traditional LBA    (b) geo-referencing based LBA

Figure 9 The graph network of traditional BA and the proposed geo-referencing based LBA.

In this paper, the locations of image can be estimated via geo-referencing, as shown in Figure 9(b), and thus a new cost function $f(x)$ is proposed to fuse the georeferencing result into LBA

$$f(x) = \frac{1}{2}\sum_{i=0}^{M}\sum_{j=0}^{N}(e_{i,j}^T w_i e_{i,j} + e_i^{G^T} w_i^G e_i^G) \tag{23}$$

where $w_i^G$ is the confidence of geo-referencing, $e_i^G = [e_x, e_y, e_z]$ are deviations between geo-referencing and visual odometry in *x, y* and *z* direction are defined as

$$e_x = |X_k^T - X_k^{T'}|$$
$$e_y = |Y_k^T - Y_k^{T'}| \qquad (24)$$
$$e_z = |Z_k^T - Z_k^{T'}|$$

where $(X_k^T, Y_k^T, Z_k^T)$ is the calculated camera position from geo-referencing, and $(X_k^{T'}, Y_k^{T'}, Z_k^{T'})$ is the camera position from visual odometry.

As the cost function $f(x)$ is nonlinear, Taylor series expansion is applied to $f(x)$ for linearization. Then, the optimization is achieved by LM (Levenberg-Marquardt).

## 5. Experiment results

### 5.1 Mars UAV image dataset

In this section, 9 simulated Mars scenes, as shown in Figure 10, were generated via Blender, a free and open-source 3D scene simulation and rendering engine [1]. The data source of DEM model and image texture includes Earth Observation data, Mars Orbital imagery and simulated Mars-like 3D terrain model. Then, based on the nine scenes, 23 UAV image sequences were generated from different trajectories. The UAV image size is 480×480 pixels, and the focal length of the camera and the sensor size are both set to 36mm, the elevation and azimuth angle of sunlight in the environment are (60°, 0°), and the light strength is 10. The Mars UAV dataset contains 40200 images in total.

---

[1] https://www.blender.org/

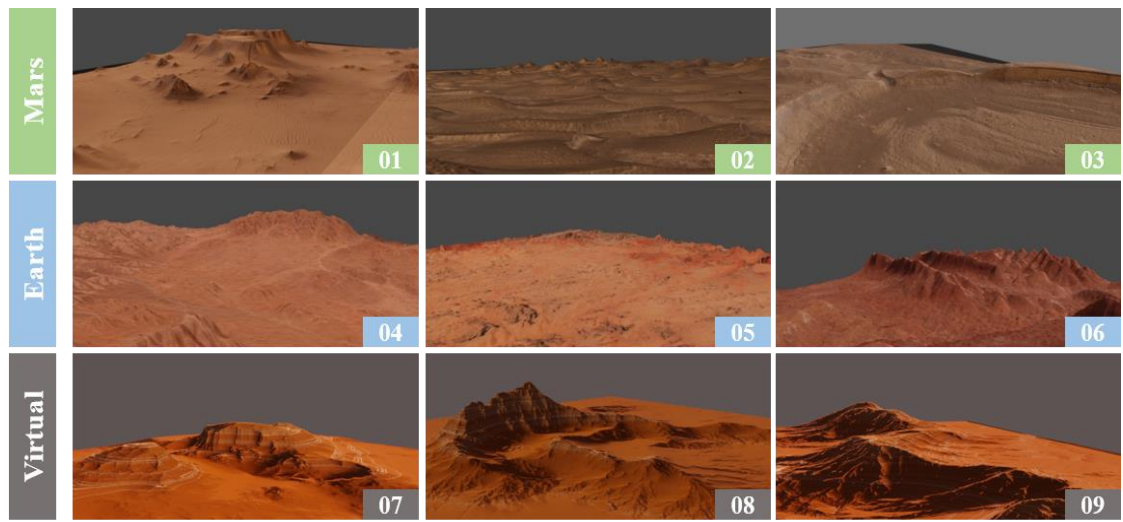

Figure 10 Nine simulated Mars scenarios generated from different sources

The simulated Mars scenes are shown in Figure 10. 01-03 scenes are generated from HiRise (High Resolution Imaging Science Experiment[2]) equipped on the Mars Reconnaissance Orbiter with ground resolution of 25cm. The three scenes include the typical Martian topography such as slope streaks, dunes and craters in Arabia Terra, Nili Patera and Breachin. 04-06 scenes are Mars-like terrain model generated from Google Map, which are scenes from Actama Desert, Jordan Desert and Chaidan Cliff. 07-09 scenes are three simulated 3D terrain models of canyons, valleys and mountains.

Statistics details of the nine Mars scenes are summarized in Table 2. As there three types of data source for Mars 3D scene simulation, our dataset represent different reflect Mars scene at different scales. The terrain model from GoogleMap which covers a large area of terrain model with rich types of topographic features. While the simulated terrain models (07-09 scene) contain detailed topographic features. The comparison of different scene scales can be seen in Figure 11. Using three Mars-like terrain models on Earth also make it possible for field trips in future. Real UAV data can be taken on Earth for accuracy assessment and algorithm performance enhancement.

Table 2 Optic Statistics summary of nine Mars scene

---

[2] https://hirise.lpl.arizona.edu/

| Scene | Terrain type | Data Sources | DEM Resolution (m/pixel) | Scene coverage (km) | Resize factor |
|---|---|---|---|---|---|
| 01 | highlands | HiRise | 1.00 | 5.9×8.5×0.5 | 10 |
| 02 | dunes | HiRise | 1.00 | 5.8×9.6×0.2 | 10 |
| 03 | crater | HiRise | 1.00 | 6.2×9.1×0.3 | 10 |
| 04 | highlands | Google Map | 8.56 | 10.3×10.3×1.5 | 20 |
| 05 | gravel | Google Map | 16.30 | 19.6×19.6×3.9 | 38 |
| 06 | mountain | Google Map | 7.64 | 9.1×9.1×0.5 | 18 |
| 07 | canyon | simulation | 0.07 | 0.6×0.6×0.07 | 1 |
| 08 | valley | simulation | 0.07 | 0.6×0.6×0.08 | 1 |
| 09 | mountain | simulation | 0.15 | 1.2×0.6×0.08 | 1 |

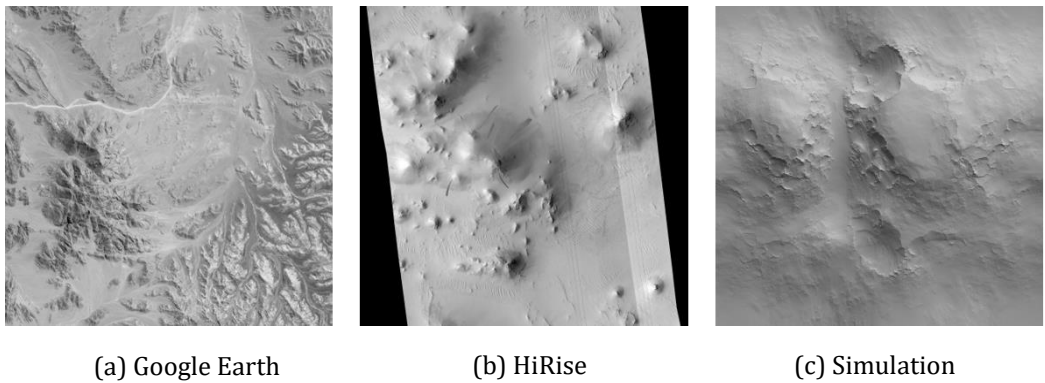

(a) Google Earth      (b) HiRise      (c) Simulation

Figure 11 Different scales in Mars scenes

To keep the size of nine scenes in consistency for UAV data generation, all the DEM models are resize to about 600×600m, together with texture image, and thus the resize factor in Table 3 vary from one scene to another. For scene 01-03, terrain texture is extracted from HiRise satellite imagery after ortho rectification. For scene 04-06, terrain texture is the surface texture extracted from GoogleMap. To simulate the color of Mars terrain, the colors of texture for 07-09 are manually changed to dark red. The interface of Blender is shown in Figure 12.

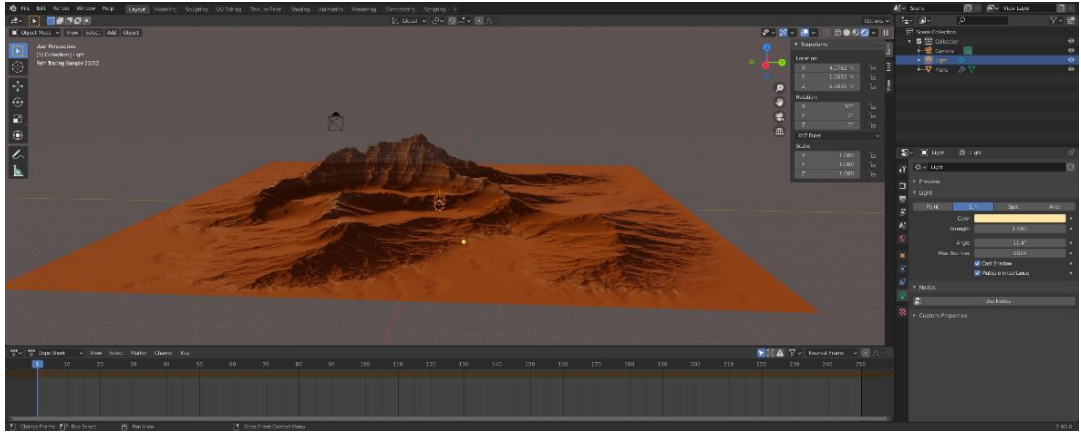

Figure 12 The interface of Blender for Mars terrain model generation

After establishing the 3D scene model of Mars, the next step is to simulate the trajectories of UAV and generate image sequence. 3 UAV trajectories, circular, forward and scanning, are generated and shown in Figure 13. The circular trajectory simulates a touch-down looking on a specific region, the forward trajectory simulates a UAV moving forward directly, and the scanning trajectory simulates the quickly mapping of a large Mars region. The whole flighting time for the first trajectories is set as 900s, and for the other two trajectories are set as 600s. The UAV viewing angle are set as nadir view to ensure a high mapping and localization precision. The camera frequency is set as 1Hz.

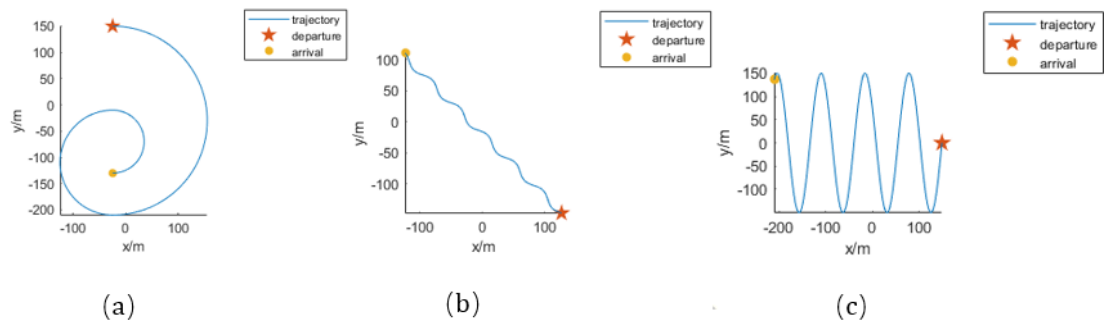

Figure 13 The three flight trajectories of the drone, (a) circular trajectory (b) forward trajectory (c) scanning trajectory (start-end)

**5.2 Image matching between UAV imagery and DEM**

In this experiment, a UAV image sequence is used as target images, while another image sequence of DEM, generated using the same trajectory, is used as reference. Since the same trajectory is used to generate the DEM and the optical image sequence, the ideal image offset is 0. Thus, the image offset between the optical images and the DEM terrain shading images can be used as metric to evaluate the image matching accuracy.

For comparison, the image matching accuracies of a feature-based method, SIFT, are calculated. Table 3 listed the matching accuracies using phase correlation and SIFT in the nine Mars scenes respectively.

Table 3 Image matching accuracies (pixel) using SIFT and our method (phase correlation)

| **Algorithms** | 01   | 02   | 03   | 04   | 05   | 06   | 07   | 08   | 09   |
| --- | --- | --- | --- | --- | --- | --- | --- | --- | --- |
| *SIFT*       | --   | 0.31 | --   | --   | 0.28 | --   | 4.39 | 0.29 | --   |
| *Our method* | 0.22 | 0.07 | 2.09 | 0.05 | 0.06 | 0.21 | 3.12 | 0.03 | 0.13 |

It can be seen in Table 3 that SIFT failed in most cases, and only succeeded in four scenes. The feature correspondence result of the grayscale images of the DEM image and the optical image in the nine scenes using SIFT is shown in Figure 14. It can be seen from Figure 14 that in most scenes, SIFT can hardly generate enough feature pairs owing to the low texture in Mars scene and large appearance variation between optical images and DEM terrain shading images.

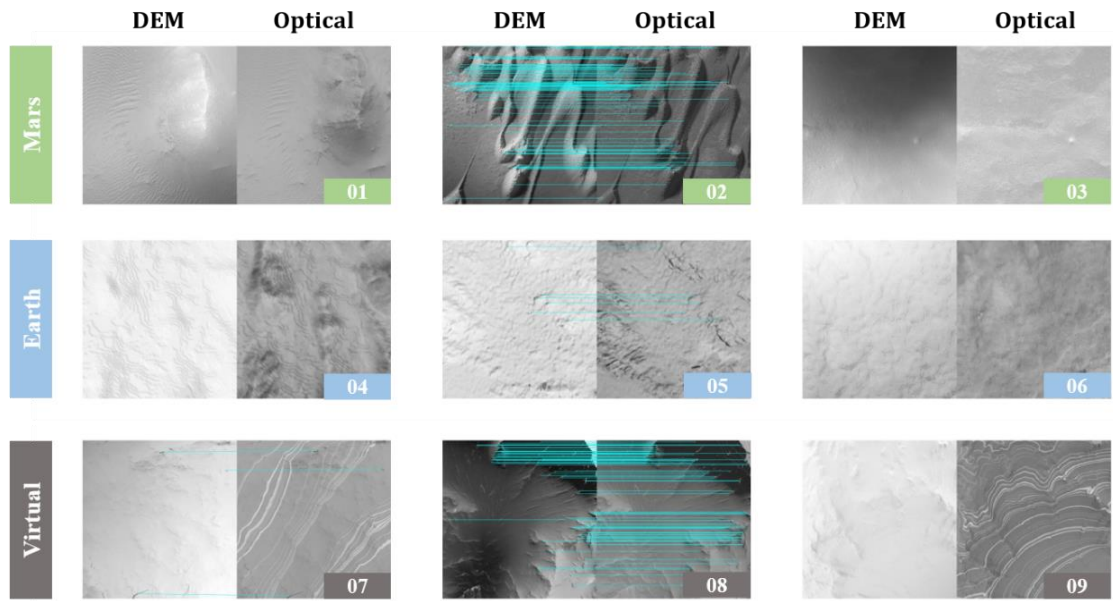

Figure 14 SIFT based image matching between optical UAV image sequence and the corresponding DEM terrain shaded images

The image matching performance of phase correlation superior to SIFT in all tested scenarios. As shown in Table 3, except for scene 3 and scene 7, the matching error of phase correlation in all scenes is less than 1 pixel, and average matching accuracy is 0.66 sub-pixel.

To further analysis the matching performance of phase correlation in nine Mars scenes, error distributions of image matching are plotted in Figure 15. It can be seen that the error distributions vary from scene to scene. For most scenes, the image offsets are distributed around 0, which can be modeled by Gaussian distribution. However, for some cases, such scene 03 and 06, the error distribution can be hardly modelled by Gaussian distribution. Figure 15 also demonstrates that although generally phase correlation achieve robust matching performance, it did fail in several cases, for example, in scene 07, for some images, the matching error can be as large as 20 pixels. This experiment paves the way for phase correlation based optical navigation approach, which fuse the geo-referencing with visual odometry for accurate and robust localization result.

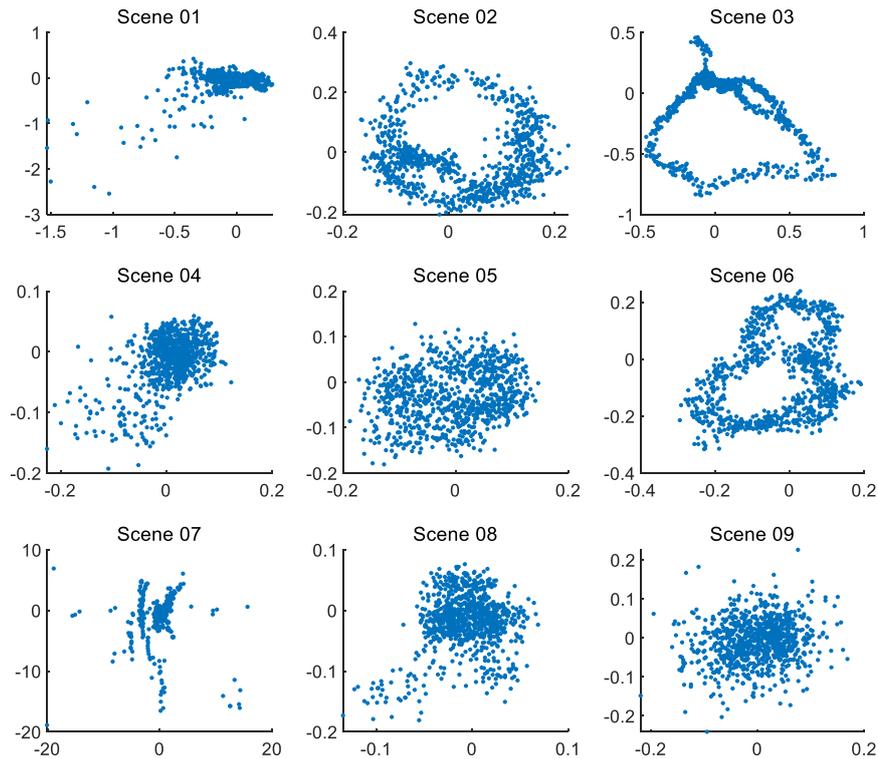

Figure 15 The error distribution of image matching accuracy using phase correlation in nine Mars scenes

**5.3 The impact of illumination condition on optical navigation**

Lighting is one of the major environmental changes on Mars which significantly alter the quality of UAV images and thus influence the navigation accuracy. The following experiments begins to explore the robustness of phase correlation to lighting conditions in the aspects of intensity, azimuth angle and elevation angle.

**5.3.1 change of illumination intensity**

This experiment will investigate the effect of lighting strength on geo-referencing results. Five UAV image sequences are generated from 3D model of scene 8 under five different illumination intensity from 0.1 to 12, as demonstrated in Figure 16.

The image sequence with the illumination intensity of 10 is set as the presumably illumination condition for terrain shading image generation, and other parameters are the same among different image sequences except for the illumination intensity. Then we test the matching accuracy of phase correlation with respect to illumination intensity using the same metric as in section 5.2.

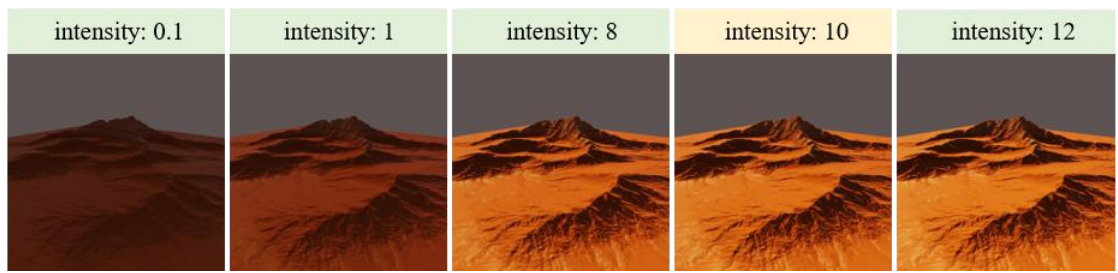

Figure 16 The appearance of scene 8 under different illumination intensity

The image matching accuracy of phase correlation under different illumination intensity are shown in Figure 17. It can be seen from Figure 17 that when the illumination intensity is 0.1, the maximum matching error is 0.129 sub-pixel, and with the lighting strength approaches 10, the error reduce to 0.047 sub-pixel. Figure 18 shows the geo-referencing error using phase correlation and has similar trend. The maximum localization error dropped from 0.032m to 0.015m with the lighting strength approaches 10.

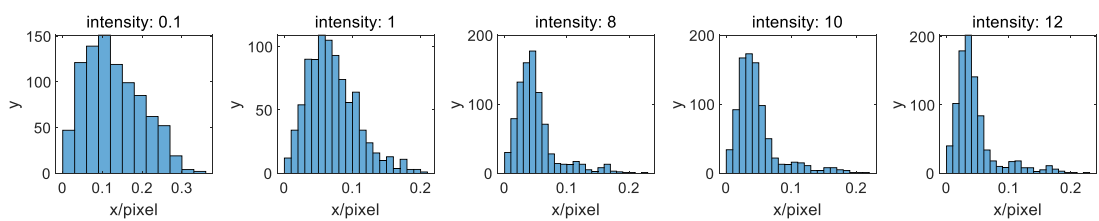

Figure 17 Pixel offsets of optical images under different illumination intensity

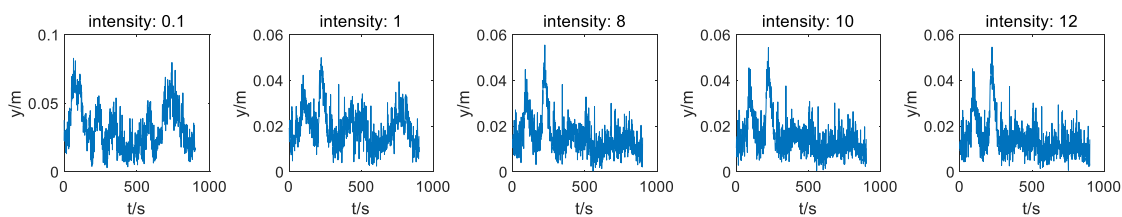

Figure 18 Absolute errors of geo-referencing under different illumination intensity

This experiment demonstrates that the accuracy of the geo-referencing is higher when the difference in light intensity between the presumably DEM image and the real optical image is smaller. However, even if the presumably light intensity is not accurate, phase correlation can still produce accurate geo-referencing result.

### 5.3.2 change of illumination azimuth

This experiment will investigate the effect of illumination azimuth on geo-referencing results. Five UAV image sequences are generated from 3D model of scene 8 under five different illumination azimuth from -15º to 45°, as demonstrated in Figure 19. The presumably azimuth angle is 0º.

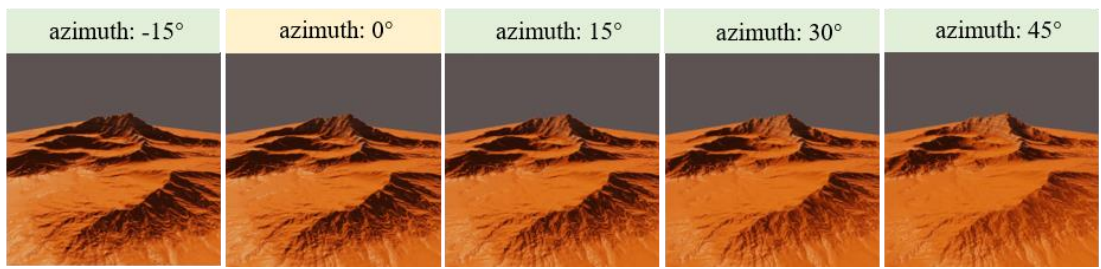

Figure 19 The appearance of scene 8 under different illumination azimuth

The accuracies of image matching and geo-referencing using phase correlation are shown in Figure 20 and Figure 21. Similar to the experiment result in section 5.3.1 that the matching error reduce to 0.047 sub-pixel when the light azimuth angle is precisely estimated.

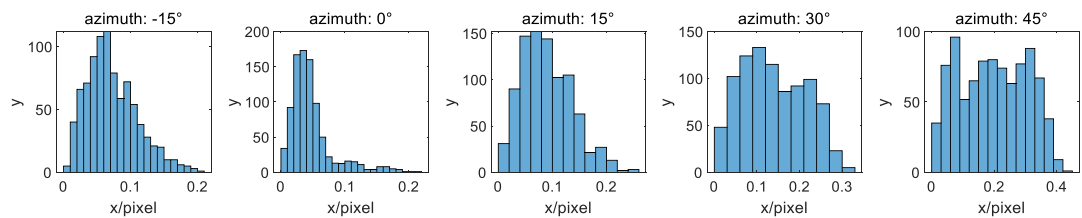

Figure 20 Pixel offsets of optical images under different illumination azimuth

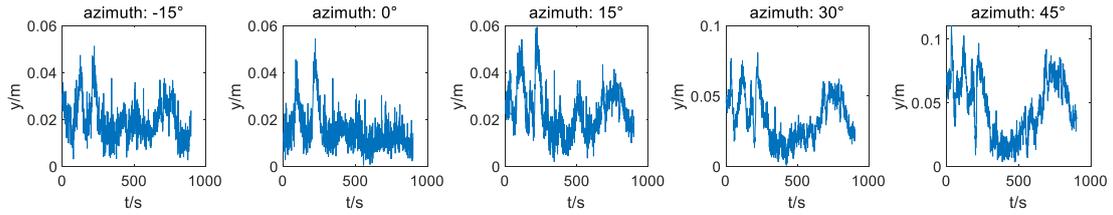

Figure 21 Absolute errors of geo-referencing under different illumination azimuth

### 5.3.3 change of illumination elevation

This experiment will investigate the effect of illumination elevation on geo-referencing results. Five UAV image sequences are generated from 3D model of scene 8 under five different illumination elevation angles from 45° to 75°, as demonstrated in Figure 22. The presumably elevation angle is 60º.

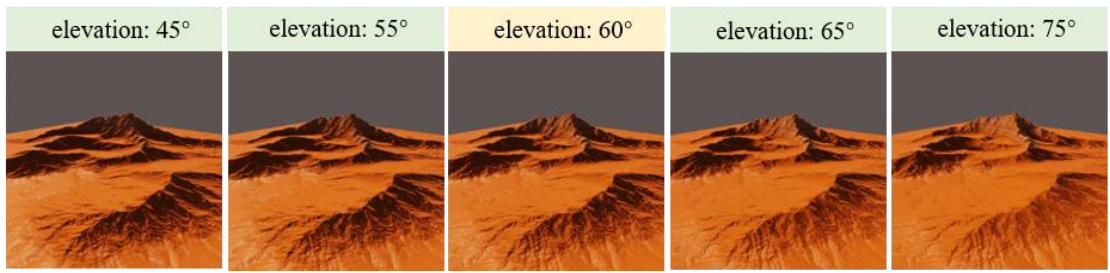

Figure 22 The appearance of scene 8 under different illumination elevation

The accuracies of image matching and geo-referencing using phase correlation under different illumination elevation are shown in Figure 23 and Figure 24. The average image matching accuracies is 0.062 sub-pixel, and the average geo-referencing accuracy is 0.089 m.

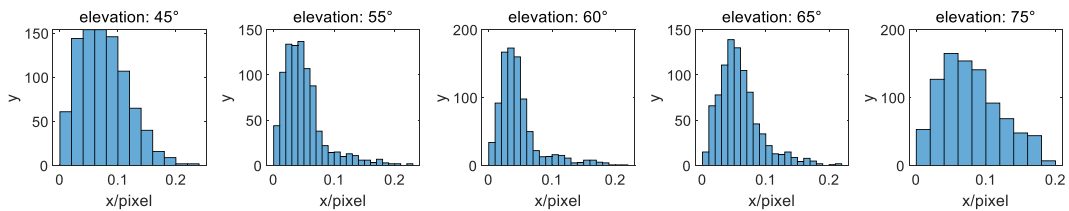

Figure 23 Pixel offsets of optical images under different illumination elevation

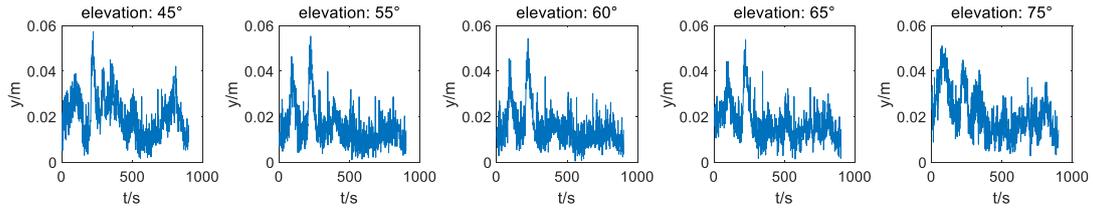

Figure 24 Absolute errors of geo-referencing under different illumination elevation

The lighting condition may not always be accurately estimated for every scheduled tasks. However, the experimental results prove that the phase correlation based geo-referencing has robustness to the uncertainty of lighting conditions. Therefore, it is quite feasible to use the geo-referencing results as the terrain constraint in the visual odometer.

**5.3 Terrain aided visual odometry experiment**

**5.3.1 The influence of the confidence of terrain constraints on the results of LBA**

The above experiment results confirm the robustness and accuracy of phase correlation based geo-referencing. In this experiment, we will further investigate the effect of the terrain constraints under different confidence values.

The terrain aided VO experiments are carried out using the UAV data of scene 8. The optical images have been generated when the light intensity is 12, the azimuth angle is 0°, and the altitude angle is 60°. A total of 6 groups of experiments are set up with the confidence level of 0, 1, 10, 100, 1000. The experiment with a confidence of 0 is equivalent to terrain constraint free. The EVO tool is used to evaluate the accuracy of calculated trajectories [3]. The absolute error of the trajectories under five different confidences are shown in Figure 25 and Figure 26.

---

[3] https://github.com/MichaelGrupp/evo

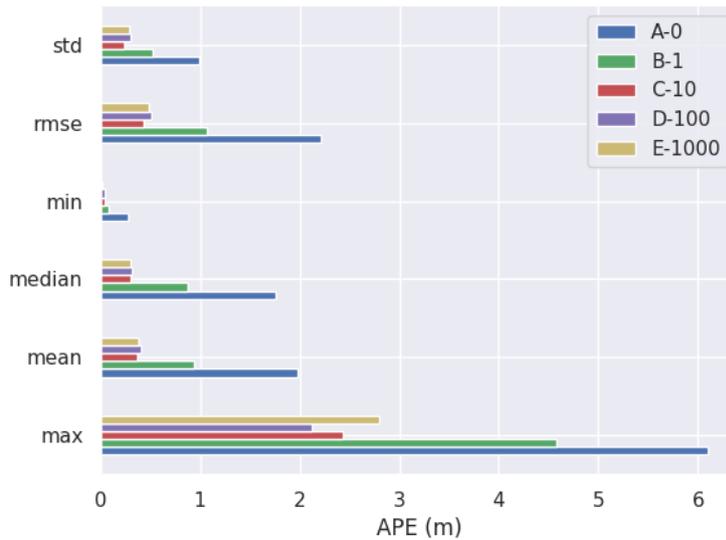

Figure 25 Absolute pose error (APE) with respect to five different confidences

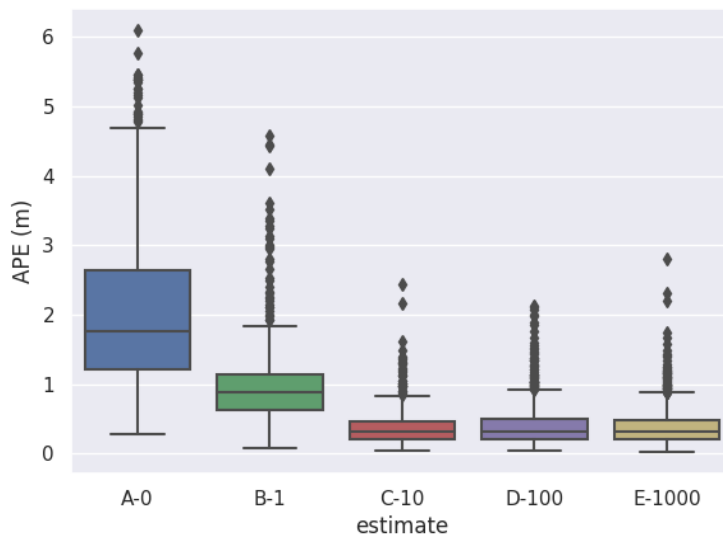

Figure 26 box plot of APE with respect to five different confidences

It can be seen from the Figure 25 that when the confidence is equal to 10, the average error, root mean square error and standard deviation are all the smallest. In addition, it can be seen that the accuracy increase significantly when the confidence level is from 0 to 10. This shows that the terrain constraint based on phase correlation has played a significant role to increase the whole navigation accuracy. however, the increase accuracy is not very obvious when the confidence is from 10 to 1000, indicating that the optimal confidence value is around 10.

Table 4 Trajectory error under five different confidences

| confidence | max/m | median/m | min/m | mean/m | rmse/m | std/m |
|---|---|---|---|---|---|---|
| 0 | 6.10 | 1.76 | 0.28 | 1.99 | 2.22 | 0.99 |
| 1 | 4.58 | 0.88 | 0.08 | 0.94 | 1.07 | 0.52 |
| 10 | 2.86 | 0.30 | **0.007** | **0.29** | **0.38** | **0.24** |
| 100 | **2.12** | 0.32 | 0.05 | 0.41 | 0.51 | 0.31 |
| 1000 | 2.80 | **0.31** | 0.02 | 0.39 | 0.48 | 0.28 |

**5.3.2 Comparison results of different flight trajectories**

This experiment will analyze the effect of UAV trajectories on the accuracies of navigation. Figure 27 shows the navigation errors of the estimated trajectories and the ground truth trajectories. The color bar on the right represents the absolute positioning error of the corresponding color. The statistic errors can be seen in Table 5.

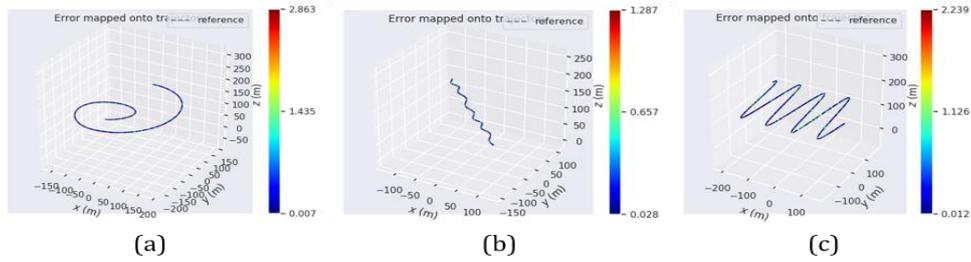

Figure 27 Errors mapped on trajectories calculated by visual odometer based on terrain constraints

As seen in Figure 27, most points in calculated trajectories are displayed in blue with low error level. Table 5 shows that compared to ORB-SLAM2 which produce 1.3 m average localization accuracy, the proposed method is able to achieve average localization accuracy of 0.3 m. ORB-SLAM2 has the lowest localization accuracy in the circular path, which may owing to the loop-closure error in circle-like trajectories, the algorithm may mistakenly consider the current position returns to the starting point while the actual trajectory is not. It is worthwhile to mention that the proposed method is not sensitive to trajectories, as the optimization is not based on loop closure, but based on geo-referencing.

Table 5 Positioning accuracy under three flight trajectories

| Trajectory | confidence | max/m | min/m | mean/m | rmse/m | std/m |
|---|---|---|---|---|---|---|
| a | ORB-SLAM | 5.61 | 0.35 | 1.92 | 2.13 | 0.93 |
|   | Our | **2.86** | **0.007** | **0.29** | **0.38** | **0.24** |
| b | ORB-SLAM | 1.93 | 0.04 | 0.55 | 0.63 | 0.30 |
|   | Our | **1.29** | **0.03** | **0.24** | **0.28** | **0.15** |
| c | ORB-SLAM | 3.15 | 0.16 | 1.15 | 1.22 | 0.41 |
|   | Our | **2.24** | **0.01** | **0.36** | **0.48** | **0.32** |

**5.3.3 Terrain aided VO experiments in all scenarios**

This experiment will investigate the performance of proposed method for all nine scenes with different terrain types and scene scales. The lighting conditions vary from scenes and can be seen in Table 6. As proved in 5.3.2, the proposed algorithm is not sensitive to flight trajectories, and thus we use circular trajectory in all 9 scenes.

Table 6 The lighting conditions of the 9 scenes in the experiment

| Scene | Light strength | Light azimuth/° | Light elevation/° |
|---|---|---|---|
| 01 | 5 | 30 | 60 |
| 02 | 2 | 0 | 60 |
| 03 | 5 | 0 | 60 |
| 04 | 5 | 30 | 75 |
| 05 | 2 | 0 | 30 |
| 06 | 5 | 0 | 30 |
| 07 | 5 | 0 | 90 |
| 08 | 10 | 0 | 60 |
| 09 | 10 | 0 | 60 |

Table 7 The positioning accuracy of the visual odometer with terrain assistance in each scene

| Scene | method | max/m | min/m | mean/m | rmse/m | std/m |
|---|---|---|---|---|---|---|
| 01 | ORB-SLAM | 5.69 | 0.08 | 1.42 | 1.60 | 0.73 |
|    | Our | **4.64** | **0.02** | **0.45** | **0.60** | **0.39** |
| 02 | ORB-SLAM | 2.16 | 0.16 | 0.87 | 0.92 | 0.29 |
|    | Our | **2.03** | **0.01** | **0.30** | **0.37** | **0.21** |
| 03 | ORB-SLAM | 1.62 | 0.04 | 0.67 | 0.71 | 0.25 |
|    | Our | **1.48** | **0.01** | **0.26** | **0.31** | **0.17** |
| 04 | ORB-SLAM | 3.52 | 0.19 | 1.02 | 1.15 | 0.54 |
|    | Our | **2.30** | **0.03** | **0.26** | **0.34** | **0.21** |
| 05 | ORB-SLAM | 1.78 | 0.01 | 0.45 | 0.52 | 0.25 |
|    | Our | **0.76** | **0.01** | **0.22** | **0.26** | **0.13** |
| 06 | ORB-SLAM | **2.02** | 0.11 | 0.89 | 0.95 | 0.31 |
|  | Our | 3.56 | **0.02** | **0.30** | **0.42** | **0.29** |
| **07** | ORB-SLAM | **3.09** | 0.21 | 0.93 | 1.00 | **0.35** |
|  | Our | 3.89 | **0.02** | **0.50** | **0.69** | 0.48 |
| **08** | ORB-SLAM | 5.61 | 0.35 | 1.92 | 2.13 | 0.93 |
|  | Our | **2.86** | **0.007** | **0.29** | **0.38** | **0.24** |
| **09** | ORB-SLAM | 5.83 | 0.77 | 2.72 | 2.89 | 0.98 |
|  | Our | **5.77** | **0.05** | **0.44** | **0.68** | **0.53** |

The navigation experimental results of the proposed method and ORB-SLAM are shown in Table 7. From the results, it can be seen that the proposed visual odometer based on phase correlation-based terrain constraints generally achieve higher accuracy than ORB-SLAM. The average localization accuracy is 0.45m while ORB-SLAM has average localization accuracy of 1.31m. It is interesting to notice that even the geo-referencing result for scene 07 is not satisfying, the final navigation accuracy of scene 07 is still under 1m. This result confirms that the local bundle adjustment is able to fuse the geo-referencing and visual odometry into a optimized result. Finally, the processing speed of the proposed method is 12fps, which ensures real-time performance.

## 6. Conclusion

In this paper, a visual SLAM method based on terrain aided geo-referencing is proposed for planetary UAV optical navigation. Based on the mathematical derivation to prove the robustness of PC to multimodal image registration, a PC based georeferencing method is proposed to automatically localize UAV image sequence on reference terrain model. The proposed optical navigation approach includes three main steps: Phase Correlation for terrain based geo-referencing, visual odometry based on frame-frame feature tracking and key frame pose refinement based on Local Bundle Adjustment. Image matching experiments between DEM and optical images and optical navigation experiments using simulated Mars UAV sequences from different terrain models have been conducted to rigorously assess the capability of the optical navigation method. The results demonstrated that compared to the state-of-the-art image matching methods, the proposed terrain aided navigation method is able to robustly recover the UAV trajectories, regardless of large appearance difference between UAV and reference DEM data.

The work presented in this paper laid theoretical foundation for terrain-aided UAV optical navigation. It can be also applied in vision-based UAV navigation on Earth when GPS signal is not accurate or unavailable. It can also be applied in the vision-based landing process on asteroid or other planet surface when terrain model is available. Future work will be carried out using the real UAV data captured from desert area which has the similar texture to planetary surface.

## Acknowledgement

The work is supported by NSFC (National Natural Science Foundation of China) 41801400.